\documentclass[sigconf]{acmart}

\usepackage{hyperref}
\usepackage{url}
\usepackage{amsmath}
\usepackage{algorithmic}
\usepackage{graphicx}
\usepackage{wrapfig}

\usepackage{tikz}
\usetikzlibrary{shapes.geometric, arrows.meta, 3d}

\tikzstyle{pipeline} = [rectangle, rounded corners, 
minimum width=4cm, 
minimum height=4cm,
text centered, 
draw=black, 
fill=red!30,
line width = 0.8mm]

\tikzstyle{pipeline2} = [
    cylinder,                     
    shape border rotate=90,        
    minimum height=4.5cm,            
    minimum width=4cm,             
    text centered,                 
    draw=black,                    
    fill=red!30,                   
    line width=0.8mm,              
    shape aspect=0.2           
]

\tikzstyle{synthesizer} = [rectangle, rounded corners, 
minimum width=1cm, 
minimum height=1cm,
text centered, 
draw=black, 
fill=yellow!20]

\tikzstyle{subsequent} = [
    rectangle, 
    rounded corners, 
    minimum width=3cm, 
    minimum height=3cm, 
    draw=black, 
    fill=yellow!30,
    line width=0.5mm,
    text height=3cm,
    text width = 2.5cm,
    text centered
]

\tikzstyle{subsequentsmall} = [
    rectangle, 
    rounded corners, 
    minimum width=2.5cm,  
    minimum height=1cm,   
    text width=2.5cm,     
    text centered,        
    draw=black, 
    fill=yellow!20,
    align=center          
]

\tikzstyle{solution} = [
    trapezium, 
    trapezium stretches=true,   
    trapezium left angle=80,    
    trapezium right angle=100,  
    minimum width=2cm,          
    minimum height=1cm,         
    text centered, 
    text width=2.5cm,           
    draw=black, 
    fill=blue!30,
]

\tikzstyle{problem} = [rectangle, 
minimum width = 3cm, 
minimum height=1cm, 
text centered, 
text width=3cm, 
draw=black, 
fill=orange!30]

\tikzstyle{data} = [diamond, 
minimum width=1.5cm, 
minimum height=1cm, 
text width=1.5cm,
text centered, 
draw=black, 
fill=green!30]

\tikzstyle{arrow} = [thick,->,>=stealth]

\usepackage{algorithm}
\AtBeginDocument{%
  }

\setcopyright{acmlicensed}
\copyrightyear{2025}
\acmYear{2025}
\acmDOI{XXXXXXX.XXXXXXX}
\acmConference[Conference acronym 'XX]{Make sure to enter the correct
  conference title from your rights confirmation email}{June 03--05,
  2018}{Woodstock, NY}
\acmISBN{978-1-4503-XXXX-X/2018/06}

\begin{document}

\title{Towards High Supervised Learning Utility Training Data Generation: Data Pruning and Column Reordering}

\author{Tung Sum Thomas Kwok}
\authornote{Both authors contributed equally to this research.}
\email{tk1018@ucla.edu}
\affiliation{%
  \institution{University of California, Los Angeles}
  \city{Los Angeles}
  \state{California}
  \country{USA}}

\author{Zeyong Zhang}
\authornotemark[1]
\email{zyong@bu.edu}
\affiliation{%
  \institution{Boston University}
  \city{Boston}
  \state{Massachusets}
  \country{USA}
}

\author{Chi-Hua Wang}
\authornotemark[1]
\email{chihuawang@ucla.edu}
\affiliation{%
  \institution{University of California, Los Angeles}
  \city{Los Angeles}
  \state{California}
  \country{USA}
}
\author{Guang Cheng}
\email{guangcheng@ucla.edu}
\affiliation{%
  \institution{University of California, Los Angeles}
  \city{Los Angeles}
  \state{California}
  \country{USA}
}

\renewcommand{\shortauthors}{Kwok et al.}

\begin{abstract}
  Tabular data synthesis for supervised learning (`SL') model training is gaining popularity in industries such as healthcare, finance, and retail. Despite the progress made in tabular data generators, models trained with synthetic data often underperform compared to those trained with original data. This low SL utility of synthetic data stems from class imbalance exaggeration and SL data relationship overlooked by tabular generator. To address these challenges, we draw inspirations from techniques in emerging data-centric artificial intelligence and elucidate Pruning and ReOrdering (`PRRO'), a novel pipeline that integrates data-centric techniques into tabular data synthesis. PRRO incorporates data pruning to guide the table generator towards observations with high signal-to-noise ratio, ensuring that the class distribution of synthetic data closely matches that of the original data. Besides, PRRO employs a column reordering algorithm to align the data modeling structure of generators with that of SL models. These two modules enable PRRO to optimize SL utility of synthetic data. Empirical experiments on 22 public datasets show that synthetic data generated using PRRO enhances predictive performance compared to data generated without PRRO. Specifically, synthetic replacement of original data yields an average improvement of 26.74\% and up to 871.46\% improvement using PRRO, while synthetic appendant to original data results with PRRO-generated data results in an average improvement of 6.13\% and up to 200.32\%. Furthermore, experiments on six highly imbalanced datasets show that PRRO enables the generator to produce synthetic data with a class distribution that resembles the original data more closely, achieving a similarity improvement of 43\%. Through PRRO, we foster a seamless integration of data synthesis to subsequent SL prediction, promoting quality and accessible data analysis. All data and codes can be accessed through the following anonymous GitHub link: \url{https://anonymous.4open.science/r/PRRO_utility}. 
\end{abstract}

\ccsdesc[300]{Theory of computation~Program reasoning}

\keywords{Training Data Generation, Supervised Learning Model Training, Synthetic Data Utility, Data Pruning, Column Reordering}


\received{1 February 2025}

\maketitle

\section{Introduction}

\begin{figure*}[htbp]
    \begin{tikzpicture}[scale=1.5]
    \draw[thick, fill=cyan!30] (-0.15,1.2) -- (1.15,1.2) -- (1.3,1.3) -- (0,1.3) -- cycle;

    \draw[thick, fill=cyan!30] (-0.15, 1.2) rectangle (1.15, -0.7); 

    \draw[thick, fill=cyan!30] (1.3,1.3) -- (1.3, -0.6) -- (1.15,-0.7) -- (1.15, 1.2) -- cycle;

    \node at (0.55, 0.3) [text width=1.5cm, align=center] {Tabular Data};
    
    
    \draw[thick, fill=orange!20] (1.7, 1.5) rectangle (3.9, 0.55); 
    \node at (2.8, 1.025) [text width=3cm, align=center] {Problem 1: Imbalanced Distribution Exaggeration};

    \draw[thick, fill=orange!20] (1.7, 0.2) rectangle (3.9, -0.75); 
    \node at (2.8, -0.275) [text width=3cm, align=center] {Problem 2: Overlook on Supervised Learning Data Relationship};

    \draw[->, thick, >=Triangle] (1.25, 0.3) -- (1.8, 1); 
    \draw[->, thick, >=Triangle] (1.25, 0.3) -- (1.8, -0.3); 

    \draw[line width=0.8mm, fill = red!30] (7.05,0.45) ellipse (0.15cm and 1.35cm);
    \fill[red!30] (4.25,-0.9) rectangle (7.05,1.8);
    \draw[line width = 0.8mm] (4.25, 1.8) -- (7.05, 1.8);
    \draw[line width = 0.8mm] (4.25, -0.9) -- (7.05, -0.9);
    \draw[line width=0.8mm, fill = red!30] (4.24,0.45) ellipse (0.15cm and 1.35cm);


    \node at (5.7, 0.33, 0) [text width=4cm, align=center] {\textbf{PRRO Pipeline}};

    \draw[line width = 0.8mm, fill=blue!30] (4.6,0.55) -- (6.6,0.55) -- (6.8,1.5) -- (4.8,1.5) -- cycle;

    \draw[line width = 0.8mm, fill=blue!30] (4.6,-0.75) -- (6.6,-0.75) -- (6.8,0.2) -- (4.8,0.2) -- cycle;

    \node at (5.85, 1.2, 0.5) [text width=2.8cm, align=center] {\textbf{Solution 1: Signal-based Data Pruning}};

    \node at (5.85, -0.1, 0.5) [text width=2.8cm, align=center] {\textbf{Solution 2: Column Conditional Reordering}};

    \draw[thick, fill=cyan!30] (7.45,1.2) -- (8.75,1.2) -- (8.9,1.3) -- (7.6,1.3) -- cycle;

    \draw[thick, fill=cyan!30] (7.45, 1.2) rectangle (8.75, -0.7); 

    \draw[thick, fill=cyan!30] (8.9,1.3) -- (8.9, -0.6) -- (8.75,-0.7) -- (8.75, 1.2) -- cycle;

    \draw[->, thick, >=Triangle] (3.9, 1) -- (4.7, 1); 

    \draw[->, thick, >=Triangle] (3.9, -0.23) -- (4.7, -0.23);

    \node at (8.15, 0.3) [text width=2cm, align=center] {Optimized Data};

    \draw[->, thick, >=Triangle] (6.7, 1.05) -- (7.65, 0.35); 

    \draw[->, thick, >=Triangle] (6.7, -0.45) -- (7.65, 0.25);     

    \draw[line width=0.5mm, fill = yellow!30] (10.3,-0.8) ellipse (1cm and 0.1cm);
    \fill[yellow!30] (9.3,-0.8) rectangle (11.3,1.6);
    \draw[line width=0.5mm] (9.3,-0.8) -- (9.3,1.6);
    \draw[line width=0.5mm] (11.3,-0.8) -- (11.3,1.6);
    \draw[line width = 0.5mm, fill = yellow!30] (10.3,1.6) ellipse (1cm and 0.1cm);

    \draw[line width=0.3mm, fill = yellow!20] (10.3,-0.3) ellipse (0.8cm and 0.05cm);
    \fill[yellow!20] (9.5,-0.3) rectangle (11.1,0.4);
    \draw[line width=0.3mm] (9.5,-0.3) -- (9.5,0.4);
    \draw[line width=0.3mm] (11.1,-0.3) -- (11.1,0.4);
    \draw[line width = 0.3mm, fill = yellow!20] (10.3,0.4) ellipse (0.8cm and 0.05cm);

    \draw[line width=0.3mm, fill = yellow!20] (10.3,0.6) ellipse (0.8cm and 0.05cm);
    \fill[yellow!20] (9.5,0.6) rectangle (11.1,1.3);
    \draw[line width=0.3mm] (9.5,0.6) -- (9.5,1.3);
    \draw[line width=0.3mm] (11.1,0.6) -- (11.1,1.3);
    \draw[line width = 0.3mm, fill = yellow!20] (10.3,1.3) ellipse (0.8cm and 0.05cm);

    \node at (10.3, -0.63) [text width=2.2cm, align=center] {\textbf{Training Data Generation}};

    \node at (10.3, -0.03) [text width=2.2cm, align=center] {Supervised Learning Model};

    \node at (10.3, 0.93) [text width=2.2cm, align=center] {Generator};

    \draw[->, thick, >=Triangle] (10.3, 0.75) -- (10.3, 0.23);   
    \draw[->, thick, >=Triangle] (8.85, 0.27) -- (9.35, 0.27);   
    



\end{tikzpicture}
\caption{Overview of the PRRO Pipeline which addresses two problems of tabular data that leads to low synthetic utility}
\label{fig: overview}
\end{figure*}
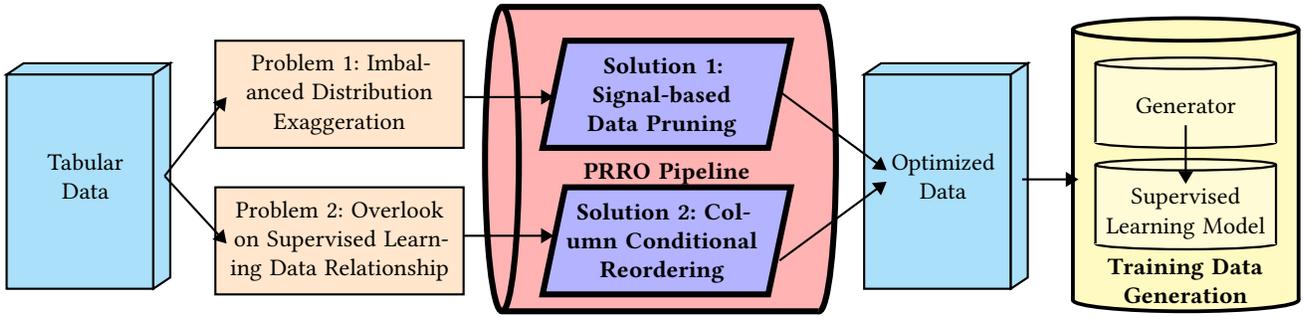
Supervised learning (`SL') training with tabular datasets is one of the most common machine learning tasks to discover patterns, learn representations and make predictions. However, real-life SL task faces constraints in data privacy and data quality, e.g. real-life datasets may contain sensitive information \cite{medical-privacy} and insufficient historical occurrences for effective analysis \cite{yebin2024synaugexploitingsyntheticdata}. Training data synthesis offers a promising solution to the above two problems by 1) replacing the original real datasets with `fake' data of similar statistical behavior and 2) appending the original real datasets with synthetic observations over the minority classes to increase exposure of the minorities \cite{data-augmentation, chang2024surveydatasynthesisapproaches, tong2024codejudgeevaluatingcodegeneration, koh2023generatingimagesmultimodallanguage, gao2024retrievalaugmentedgenerationlargelanguage}. The development of the training data synthesis area has been immensely accelerated by the speedy evolution of generative AIs in terms of model training cost and performance \cite{sengar2024generativeartificialintelligencesystematic, ye2024generativeaivisualizationstate, deepseekai2024deepseekv3technicalreport} as well as practitioners' realization of value after applications in business, medicine and marketing industry \cite{synthetic-use-case-1, synthetic-use-case-2, synthetic-use-case-3, synthetic-use-case-4,synthetic-use-case-5}. \\
\\
As generative models continue to push the boundaries of data synthesis, it becomes crucial to examine not just the fidelity of the generated data, but also the utility: how effectively it can be used in downstream machine learning tasks. This evaluation metric plays a particularly important role in tabular data for conducting subsequent data analysis work, despite being underexplored compared to synthetic fidelity in many works \cite{yuan2024realfakeeffectivetrainingdata, derec}. Combined with works showing that high fidelity synthetic data does not necessarily contribute to high utility \cite{xu2024utilitytheorysyntheticdata}, there is the urgent industrial need to explore major contributing factors that govern synthetic data utility. Specifically, high SL utility of synthetic data is primarily hurdled by the following factors: (1) Class imbalance is extremized (Fig. \ref{fig: result-summary}) due to distribution smoothening in machine learning models \cite{smoothening}, which treats low-occurrence class as outliers / noise \cite{nuggehalli2024directdeepactivelearning} and exaggerating the imbalance (Table \ref{tab: pruning-results}). When studying synthetic data of highly imbalanced dataset, which undergoes two consecutive smoothening steps in the generator and SL model, the exaggeration problem is especially severe. The urgency of mitigating this exaggeration is accelerated along the increasing prevalence of synthetic data.; (2) SL model's key focus on predictor-feature conditional relationship is overlooked by the table generator. While the latter model type regards data relationships over the entire table equally \cite{borisov2023language, sdv}, the former has a well-established statistical assumption of predictor conditional on features \cite{downstream-assumptions}, which generator weighs less on. Bridging the two conditional relationship assumptions is needed to allow smooth translation between output from upstream generators and input to downstream SL models. \\
\\
To address the aforementioned challenges, in this paper, we introduce a novel PRuning and ReOrdering PRRO pipeline, a pre-synthesis-processing pipeline that guides the generator to generate high SL utility synthetic data by complying to common SL techniques and assumptions.  The initial phase of PRRO pipeline aims to narrow down the focus of data modeling, by emphasizing on the group of data that we seek to study and pruning observations that have low correlation to our interest group. This balances the dataset class distribution, hence reducing the extent of exaggeration during synthesis. Following this, PRRO employs a reordering module to standardize optimal order of table features for data synthesis. This guides the generator to establish a tabular conditional relationship comparable to that assumed under downstream SL models, hence preventing drop of SL utility due to overlooking of key relationships. All in all, this data pre-synthesis-processing approach bridges the difference in assumptions between generators and downstream SL models, streamlining the 'generate and predict' pipeline. \\
\\
In our study, we conduct extensive experiments across two application scenarios of synthetic data, namely data replacement and appendant, using datasets from a large collection of prediction tasks in previous studies \cite{abalone-use, credit-use, titanic-use}. This collection includes 22 different types of classification tasks across identifying between age of creatures, financing behaviors of users, academic performance of students, and advertisement-clicking records of online browses. Our study stands out for its coverage on evaluating SL utility of synthetic data under different applications and class distribution variability, setting a new benchmark for SL utility evaluation when compared to previous work \cite{borisov2023language, solatorio2023realtabformergeneratingrealisticrelational, nguyen2024generatingrealistictabulardata}. The empirical results highlight improvement in overall SL utility when original data is appended with synthetic data generated through the pipeline. To summarize, our main contributions are as follows: \begin{enumerate}
    \item In Sec. \ref{sec: preliminary}, formally defining supervised learning utility and hence synthetic data utility for synthetic data evaluation specifically;
    \item In Sec. \ref{sec: pipeline}, identifying two major contributing factors to low SL utility of synthetic data, namely generator exaggerating class imbalance and overlooking key SL data relationship, hence introducing the PRRO pre-processing pipeline, which prunes and reorders tabular data to address the two factors;
    \item In Sec. \ref{sec: experiment}, showing PRRO implementation contributes to consistent SL utility improvement over experiments on a broad range of datasets. 
\end{enumerate}

\section{Preliminary} \label{sec: preliminary}

\subsection*{Supervised Learning Utility of Synthetic Training Data}

Given a labeled training dataset $D_T=\left\{\mathbf{x}_{T, i}, y_{T, i}\right\}_{i=1}^{n_T}$, where $\mathbf{x}_{T, i}$ denotes the features for observation $i$ and $y_{T, i}$ the corresponding label, our goal is to train a supervised learning (SL) model $h$ to predict $y$ from $\mathbf{x}$. Formally, we seek a model $h(y \mid \mathbf{x})$ that minimizes the expected loss over $D_T$ :
\begin{equation}
l\left(h ; D_T\right)=\frac{1}{n_T} \sum_{i=1}^{n_T} l\left(h ; \mathbf{x}_{T, i}, y_{T, i}\right)
\end{equation}
Given a validation dataset $D_V=\left\{\mathbf{x}_{V, i}, \boldsymbol{y}_{V, i}\right\}_{i=1}^{n_V}$, the performance of a supervised learning model $h$ is assessed via the \textbf{validation loss},
\begin{equation}
l\left(h ; D_V\right)=\frac{1}{n_V} \sum_{i=1}^{n_V} l\left(h ; \mathbf{x}_{V, i}, y_{V, i}\right)
\end{equation}
In many classification settings, for instance, $y$ may be a binary class label (0 or 1), and a common choice for $l(\cdot)$ is the cross-entropy loss: $l(h ; \mathbf{x}, y)=-[y \log \hat{y}+(1-y) \log (1-\hat{y})]$
where $\hat{\boldsymbol{y}}=h(\mathbf{x})$ is the predicted probability of the positive class. The validation loss then captures the average divergence between the predicted and true labels on the validation set.

\begin{definition}[Supervised Learning Utility]\label{def: supervised-learning-utility} \emph{Supervised learning utility} is defined as the difference in validation loss between a \emph{target model} $h_A$ and a \emph{baseline model} $h_B$ on the same validation dataset $D_V$ :
\begin{equation}
U\left(h_A ; h_B, D_V\right)=l\left(h_A ; D_V\right)-l\left(h_B ; D_V\right)
\end{equation}
\end{definition}

Definition \ref{def: supervised-learning-utility} can be tailored to measure \emph{synthetic data utility} by comparing a model trained on synthetic data to one trained on the original data. Specifically, let $D_T$ be partitioned into a \emph{generator-training} set $D_{S T}= \left\{\left(\mathbf{x}_{S T, i}, y_{S T, i}\right)\right\}_{i=1}^{k_{S T}}$ and a \emph{holdout} set $D_H=\left\{\left(\mathbf{x}_{H, i}, y_{H, i}\right)\right\}_{i=1}^{k_H}$ to prevent the generator from over-fitting all training data points \cite{holdoutintro, train-holdout}. A data generator $\mathcal{G}_A$ is then fitted using $D_{ST}$ to produce a synthetic dataset $\tilde{D}_A$, i.e.
\begin{equation}
\mathcal{G}_A: D_{S T} \longrightarrow \tilde{D}_A
\end{equation}
Afterward, one can train separate predictive models $h_D$ on either the original dataset $D_T$ or the synthetic dataset $\tilde{D}_A$.

\begin{definition}[Synthetic Data Utility]\label{def: synthetic-data-utility} \emph{Synthetic data utility} is the supervised learning utility (Definition \ref{def: supervised-learning-utility}) when the target model is trained on synthetic data $\tilde{D}_A$ and the baseline model is trained on the original data $D_T$ :
\begin{equation}
U\left(h_{\tilde{D}_A} ; h_{D_T}, D_V\right)=l\left(h_{\tilde{D}_A} ; D_V\right)-l\left(h_{D_T} ; D_V\right)
\end{equation}
\end{definition}

This measure captures how well the synthetic dataset $\tilde{D}_A$ \emph{replaces} the original data $D_T$. To measure how synthetic data may \emph{augment} the original data, one can define a combined dataset $\hat{D}_A=D_T \cup \tilde{D}_A$. The synthetic data utility in this appendant scenario is then
\begin{equation}
U\left(h_{\hat{D}_A} ; h_{D_T}, D_V\right)=l\left(h_{\hat{D}_A} ; D_V\right)-l\left(h_{D_T} ; D_V\right) .
\end{equation}

For a concrete example in a classification task, this utility would reflect how adding (or replacing with) synthetic data affects metrics such as validation accuracy or cross-entropy loss, indicating whether the synthetic dataset enhances or degrades the predictive performance of the trained model.

\section{Proposed Approach - Pruning and ReOrdering PRRO Pipeline} \label{sec: pipeline}
We propose the Pruning and ReOrdering (PRRO) pre-processing pipeline, to address two factors observed that lead to degradation of SL utility for synthetic data: 1) Imbalanced class distribution is leveraged by generators, and 2) SL data relationship is overlooked in existing synthetic data research, which misguides generator from capturing the correct column relationship. The PRRO pipeline modifies the original dataset into a format $D^{*}$ that bridges the gap between generators and SL models for data synthesis. After synthesis, the resulting dataset is in a post-pipeline format and must be inversely transformed to its original structure for consistency. 



\subsection{Signal-based Data Pruning to mitigate imbalance class exaggeration} \label{sub: signal-subgrouping}

Generators, like most machine learning models, exaggerate on imbalanced dataset and heavily favor the majority class due to distribution smoothening. Signals from minority class of interest are dominated by the majority class which contains a lower signal-to-noise ratio. We observe generators generating \textit{only} majority class data points in some highly imbalanced datasets (Table \ref{tab: pruning-results}), indicating a more extreme class distribution in synthetic data over original data. With the synthetic class distribution being more imbalanced, SL models struggle further in capturing signals from the minority class, rendering low SL utility for the dataset. Techniques such as oversampling and SMOTE \cite{stocksieker2023generalizedoversamplinglearningimbalanced, Chawla_2002} are proposed to balance the imbalanced distribution, but \cite{hassanat2022stopoversamplingclassimbalance} shows that oversampling methods are unreliable to learn from class imbalanced data because of their tendency to generate the 'majority of minority', instead of considering the distribution of minority class, e.g. SMOTE synthesizes data by interpolating existing minority group, but interpolation struggles in high-dimensional spaces \cite{elor2022smotesmote} when the distance between points becomes less meaningful due to "curse of dimensionality" \cite{peng2024interpretingcursedimensionalitydistance}. In turn, works on undersampling, specifically data pruning such as DRoP carry analogies with distributionally robust optimization methods to address class/group bias \cite{vysogorets2024dropdistributionallyrobustpruning}. However, while the DRoP method is respected to take into account class-conditional independence, it does not hold strongly for more sophisticated group-wise datasets when the correlation across different observations is different. Therefore, we propose a data pruning method (Sec. \ref{sub: signal-subgrouping}) that further considers different class-conditional relationships across observations, hence focusing on data having high correlation with our group of interest. \\
\\
To address the exaggeration of imbalanced dataset, we propose the signal-based data pruning module to balance the class distribution of original data to mitigate the exaggeration effect of generator. Given the training dataset $D_{T}$, we extract the subgroup containing observations labeled with our interested class (denoted as $D'\subset D$) and leave the remaining non-interested class observations (denoted as $\not D'\subset D$). By iteratively going through each data point in the non-interested class subgroup, we identify and extract observations with considerable correlation with the interested subgroup data: $D'^{*} = \{d\in \not D'|\exists d' \in D':corr(d, d')>\tau\}$. These observations with non-interested class but containing high relevancy to our study interest are then concatenated with the original interested subgroup to form the desired signal group: $D^{*}=D'\cup D'^{*}$. This signal group with higher signal-to-noise ratio serves as exemplar and guides the generator towards studying specific traits of our class of interest.  \cite{yuan2019signaltonoiseratiorobustdistance}.\\
\\
This data pruning module is specifically designed to assist in data analytics tasks over imbalanced datasets. We illustrate the method with a binary classification setting, where the group of interest is the positive observations. Machine learning models tend to focus and bias towards the majority class, e.g. in the modeling of CTR datasets where the average industrial clickthrough rate is about $0.57\%$ \cite{averageclickingrate}, both the synthetic and downstream predictive models are incentivized to model the majority class that accounts for $99.43\%$ in terms of maximizing overall accuracy. This majority bias, which also extends to state-of-the-art models such as LLM \cite{gallegos2024biasfairnesslargelanguage}, can be problematic when the model generates / predicts no positive observations at all, as the majority negative class is not of our interest, hence contrasting the research goal. The proposed data pruning method increases the density of the informative observations in the subdataset to be studied, allowing the generator to lean towards studying and generating the minority class. \\
\\
Algo. \ref{algo: signal-subgrouping} shows the algorithm for the Signal-based Data Pruning module in binary classification case. The algorithm firstly separates the dataset into the group of interested and non-interested subgroups. Each observation from the non-interested class will undergo a correlation measurement with every observation from the interested class, and will be classified as the signal-containing subgroup if this observation exhibits significant correlation with any interested class observations. The correlation measurement in our experiment involves a numerical measurement using the Spearman correlation, whose motivation will be further discussed in Sec. \ref{subsub: pipeline implementation}. The algorithm focuses on pruning the observations that 1) contain a class that researchers are not interested in learning, and 2) have negligible contribution in assisting researchers to learn about the class of interest, which summarizes into noisy observations with limited contribution to the SL task. 

\begin{algorithm}[t]
\caption{Signal-based Data Pruning Algorithm}\label{algo: signal-subgrouping}
\begin{algorithmic}[1]
\STATE \textbf{Input:} The training dataset $D_{T}=\{(\textbf{x}_{T,i},y_{T,i})\}^{k_{T}}_{i=1}$, where $D'=\{(\textbf{x}_{T,i},y_{T,i})|y_{T,i}=1\}\subset D$ and $D''=\{(\textbf{x}_{T,i},y_{T,i})|y_{T,i}=0\}\subset D$.
\STATE \textbf{Output:} The signal-containing subgroup containing interested class observations and other non-interested class observations with high correlation to the interested class observations: $D^{*}=D' \cup \{(\textbf{x}_{T,i},y_{T,i})\in D''|\exists x_{k}\in D': \text{corr}(\textbf{x}_{T,i},\textbf{x}_{k})>\tau\}$.
\STATE Initialize $\tau$, $D^{*} \gets D'$;
\FOR{each $d0 \in D''$}
    \FOR{each $d1 \in D'$}
        \IF{$\text{corr}(d0, d1) > \tau$}
            \STATE $D^{*} \gets D^{*} \cup d$;
            \STATE \textbf{break}
        \ENDIF
    \ENDFOR
\ENDFOR
\end{algorithmic} 
\end{algorithm}

\subsection{Column Conditional ReOrdering to replicate supervised learning data relationship} \label{sub: optimal-feature} 
Generators often overlook critical data relationships required by SL models. This is because SL models focus specifically on the predictor-features conditional relationship, whereas other machine learning paradigms, such as unsupervised learning and reinforcement learning, examine broader relationships across features and environmental states \cite{slvsusl}. While synthesizers typically prioritize the overall tabular structure to meet the needs of various machine learning tasks, they tend to neglect the specific predictor-feature conditional relationship that is vital for supervised learning models. As a result, synthetic data does not sufficiently exhibit the SL-focused relationship. Generators study the inter-column conditional relationships either by the overall table cross-feature relationship \cite{tabddpm, ctgan} or the column-by-column sequential relationship \cite{sdv, borisov2023language}. Specifically, state-of-the-art generators such as LLM are brittle to ordering of inputs, even if the ordering does not alter the underlying task \cite{chen2024premiseordermattersreasoning, anonymous2024eliminating}. In contrast, SL models analyze the conditional relationship between the predictor and features while assuming minimal conditional dependence among features \cite{downstream-assumptions}. When such dependencies are non-negligible, additional preprocessing techniques are applied to mitigate multicollinearity and prevent overfitting \cite{multicollinearity-in-regression, gómez2024enlargingsampleaddressmulticollinearity, shlens2014tutorialprincipalcomponentanalysis}. Consequently, the generator does not adequately account for the conditional relationship between the predictor and features when generating SL-compatible data. To bridge this gap, we propose the Column Conditional Reordering method (Sec. \ref{sub: optimal-feature}) by replicating the predictor-features conditional relationships of SL models using generators' sequential-column-based assumption. A closer resemblance of conditional relationship modeling between the two models guides the data generator to 'think and learn' like the SL model and to generate data accordingly.\\
\\
Inspired by confidence learning method that trains SL model by prioritizing observations with higher prediction confidence, i.e. more obvious patterns \cite{northcutt2022confidentlearningestimatinguncertainty}, we propose a Column Conditional ReOrdering module to guide the generator to model data in closer resemblance as SL model. SL model has a well-established assumption on studying the conditional distribution of predictor given features. While generators are not originally designed to focus on this specific conditional relationship, LLM-based generators do model data conditionally under a sequential-column-based assumption. By rearranging the column features, the sequential-column-based assumption of generators can be leveraged to mimic the conditional relationship between predictor and features. Subsequent SL model training can be enhanced by utilizing this synthetic dataset generated under the closely-aligned conditional relationship. \\
\\
We conduct the experiment by comparing between the synthetic data utility of synthetic data generated under different predictor-features conditional relationship. Our experiment does not emphasize on the relationship between features, due to the assumption that feature data for SL model training should have a negligibly low multicollinearity. Under this assumption, the conditional distribution of a feature given previous features equates to the marginal distribution of that feature itself: $f(B|A)=f(B) \Longleftrightarrow A \perp B$. Therefore, the only conditional distribution to be considered is then between the features and the predictor, where the predictor should be placed after the features to mimic a conditional relationship of predictor given the features assumed from SL models. The first batch of data is inputted into the generator accordingly: $\{\textbf{x}_{T},y_{T}\}$ for generation. To show the importance of assumption compliance, we also generate another batch of data by placing the predictor before the features as a control group: $\{y_{T}, \textbf{x}_{T}\}$, simulating the synthesis under the features given predictor conditional relationship. \\
\\
Algo. \ref{algo: optimal-feature-simp} standardizes the Column Conditional ReOrdering method by formulating predictor-feature conditional relationship in generator to align with the respective data modeling focus of SL models. Regardless of where the predictor is located initially in the dataset, the algorithm ensures that the predictor would be placed at the last column of the dataset to guide the generator to model the predictor-feature conditional correlation. Given the no-multicollinearity assumption, the method does not reorder the features sequence, despite an extension of the Column Conditional ReOrdering module will be discussed in Sec. \ref{sec: future work} which further takes into account multicollinearity problem over realistic datasets. Moreover, the algorithm adds a control group testing by arbitrarily placing the predictor column of all datasets to the first column. Both synthesis will be carried out under the same generator setting detailed in Sec. \ref{subsub: model specifications} to mitigate potential modeling variance. This guides the generator in modeling in different conditional distribution assumptions between features and predictor, hence validating the significance to have the generator complying with the requirement in SL models. 

\begin{algorithm}[t]
\caption{Formulation of predictor-feature conditional relationship in generator}\label{algo: optimal-feature-simp}
\begin{algorithmic}[1]
\STATE \textbf{Input:} A training and a validation dataset: $D_{T/V}=\{(\mathbf{x}_{T/V,i},y_{T/V,i})\}_{i=1}^{k_{T/V}}$
\STATE \textbf{Output:} Synthetic data utility difference when data is generated in compliance with a supervised learning model versus when it is not.

\STATE Initialize generator $\mathcal{G}_{A}$ predictive model $h$
\STATE $X \gets \mathbf{x}_{T}$;
\STATE $df_{1} \gets \text{concat}(X, y)$;
\STATE $df_{2} \gets \text{concat}(y, X)$;

\STATE Train two $\mathcal{G}_{A}$ using $df_1$ and $df_2$ separately;
\STATE Synthesize $\tilde{df}_1$, $\tilde{df}_2$;

\STATE Initialize predictive models $h_1$, $h_2$;
\STATE Train $h_1$ and $h_2$ with $\tilde{df}_1$ and $\tilde{df}_2$ respectively;
\STATE Compare $error(h_1(\mathbf{x}_V) - y_V)$ and $error(h_2(\mathbf{x}_V) - y_V)$;
\end{algorithmic}
\end{algorithm}

\section{Experiment} \label{sec: experiment}
We are now ready to experiment on the PRRO pipeline, by pruning observations with low signal-to-noise ratio and comparing between synthetic data utility between placing the predictor before and after the features in the dataset. The experiment is conducted over 22 binary classification datasets with different levels of class distribution. \vspace{-3mm}  

\subsection{Experimental Setups}
\subsubsection*{\textbf{Datasets Pre-processing}} 
We conduct our experiments over a collection of real-world datasets, listed in Table \ref{tab: data-summary}, commonly used in previous works for tabular classification tasks \cite{titanic-use, abalone-use, derec, credit-use}. These datasets differ with respect to the number of observations, ranging from 96 up to 134,868 observations. There are also variations in the number of features each dataset contains, ranging from 2 to 14 numerical variables and 1 to 19 categorical variables. To pre-process the datasets, we take the following three steps: \textbf{1) Each dataset is divided into a training, holdout and validation dataset} according to Sec. \ref{sec: preliminary} in a 40\%:40\%:20\% ratio \cite{train-holdout, holdoutintro}. This is to check if the synthetic data overfits the training dataset, as it implies excessive information leakage during data synthesis \cite{zhang2020privacyalldemystifyvulnerability}; \textbf{2) We subset the CTR dataset from \cite{ctr_dataset} into multiple subdatasets to conduct a more in-depth study on synthetic data utility of extremely imbalanced datasets} (with positive rate below 5\%). The dataset is separated according to the types of advertisement channel and advertisement contents to mimic advertisement clicking behavior under different electronic devices and channels, so that each subdataset is considered independent and aligns with authorized marketing data guideline \cite{admap}, and \textbf{3) We obtain all non-CTR datasets directly from HuggingFace API, without any additional manual pre-processing} from the team to isolate the usage of pipeline for efficacy testing. Details of the datasets are included in Appendix \ref{tab: data-summary}. 

\subsubsection*{\textbf{Pipeline Implementation}} \label{subsub: pipeline implementation}
The two modules in PRRO pipeline are implemented accordingly as follows: \textbf{1) The implementation of the Signal-based Data Pruning module (Algo. \ref{algo: signal-subgrouping}) involves defining correlations between observations}. Since most datasets contain both continuous and categorical variables, the Spearman rank correlation coefficient is used to ensure broad applicability given its ability in handling both ordinal and continuous data \cite{correlation}. In our experiment, an observation from the non-interest subgroup is considered correlated with the interest subgroup if its Spearman correlation coefficient exceeds 0.3 with any observation in the interest subgroup. This is because a Spearman correlation of 0.3 represents a weak positive monotonic relationship \cite{cohen1988statistical}, where this weak correlation still holds useful information in real-world data in areas such as marketing and healthcare \cite{norman2010likert, miller2005applying}; \textbf{2) The Column Conditional ReOrdering module compares two batches of standardized datasets, both containing the same data}. In the first batch, the predictor is manually moved to the last column (if it is not already there), while in the second batch, the predictor is moved to the first column (if it is not already there). This approach mimics modeling the data in terms of the conditional relationship between predictors and features versus features and predictors. By adjusting the position of the predictor, we account for variations in how predictors are positioned across datasets, ensuring standardized comparisons. This results in three distinct experimental settings: (1) the model trained on the original data, (2) synthetic data following the SL modeling structure: $\tilde{D} = (\tilde{y}, \tilde{X})$, and (3) synthetic data that does not follow the SL modeling structure: $\tilde{D} = (\tilde{X}, \tilde{y})$.

\subsubsection*{\textbf{Model Specifications}} \label{subsub: model specifications}
There are two data modeling stages on generators and classification models. All datasets are experimented under the same generator and classification models of the following specifications: \textbf{1) Regarding the data generator, this experiment utilizes the \texttt{REaLTabFormer} package from \cite{solatorio2023realtabformergeneratingrealisticrelational} as the primary generator}, with regard to LLM-based generators' outperformance from other works \cite{borisov2023language, solatorio2023realtabformergeneratingrealisticrelational, nguyen2024generatingrealistictabulardata, derec}. While the online available \texttt{REaLTabFormer} can be installed directly, the model is initially designed for a fine-tuned but out-dated GPT-2 model. It is then rewritten to allow generalized usage of LLMs available in HuggingFace along with some bugs fixed to account for the generative power of SOTA LLMs (updated package will be released in GitHub in the future after completion in generalizing multi-table use cases). In this experiment, we utilize the lightweight Llama-3.2-1b \cite{llama32} as baseline because its small size allows usage from most local computers for reproducibility. For each synthesis, the LLM is fine-tuned with 500 epochs and 10 batches per epoch; \textbf{2) Regarding the downstream SL classification models, we utilize multiple classification models to study if the pipeline is model-consistent}. The models include Logistic Regression \cite{lr}, four Tree-based Classifiers: Decision Tree, Random Forest, Gradient Boosting and Extreme Gradient Boosting \cite{decisiontree, featureimportance, friedman2001greedy, Chen_2016}, three Naive Bayes models: Gaussian, Complement and Multinomial Naive Bayes \cite{duda2000pattern, wei2005complement, mccallum1998improvements}, and two neural network models: DNN and DeepFM \cite{lecun2015deep, guo2017deepfmfactorizationmachinebasedneural}. Aside from the neural network models that are written from PyTorch \cite{paszke2019pytorchimperativestylehighperformance}, with ReLU activation function and 3 layers of 256 nodes in the deep component, other models are called as default class from the Sci-kit Learn package \cite{scikit-learn}.

\subsubsection*{\textbf{Evaluation Metric}}
Evaluation of synthetic data utility is conducted on how the synthetic data serves as the replacement and appendant to the original data (Def. \ref{def: synthetic-data-utility}). Given the experiment is binary-classification focused, we evaluate the synthetic data utility based on classification accuracy. However, measurement based on overall accuracy can be problematic when measuring highly imbalanced datasets, for instance, if making all predictions to be the majority class yields a high prediction accuracy. \textbf{Therefore, we further evaluate synthetic data utility with respect to precision, recall, F1 score and AUC score}, by replacing the loss function in Def. \ref{def: synthetic-data-utility} with the each respective metric function. We denote the metric set $\mathcal{M}$ containing the above metrics that evaluates dataset $D$ as $\mathcal{M}_{D}=\{P_{D}, R_{D}, F1_{D},AUC_{D}\}$, where 
\begin{align*}
    P_{D} &= \mathbb{P}_{(\mathbf{x},y)\in D_{V}}(y=1|h_{D}(\mathbf{x})=1),\\
    R_{D} & = \mathbb{P}_{(\mathbf{x},y)\in D_{V}}(h_{D}(\mathbf{x})=1|y=1),\\ 
    F1_{D} &= 2 \times \frac{P_{D}\times R_{D}}{P_{D}+R_{D}}, \\
    AUC_{D} &= \mathbb{P}_{(\mathbf{x}_{1}, y_{V,1}), (\mathbf{x}_{2}, y_{2})\in D_{V}}(h_{D}(\mathbf{x}_{1})>h_{D}(\mathbf{x}_{2})|y_{1} = 1, y_{2} = 0).
\end{align*}

\subsection{Results} 
\begin{figure*}[htbp]  
\centering
\includegraphics[width=\textwidth]{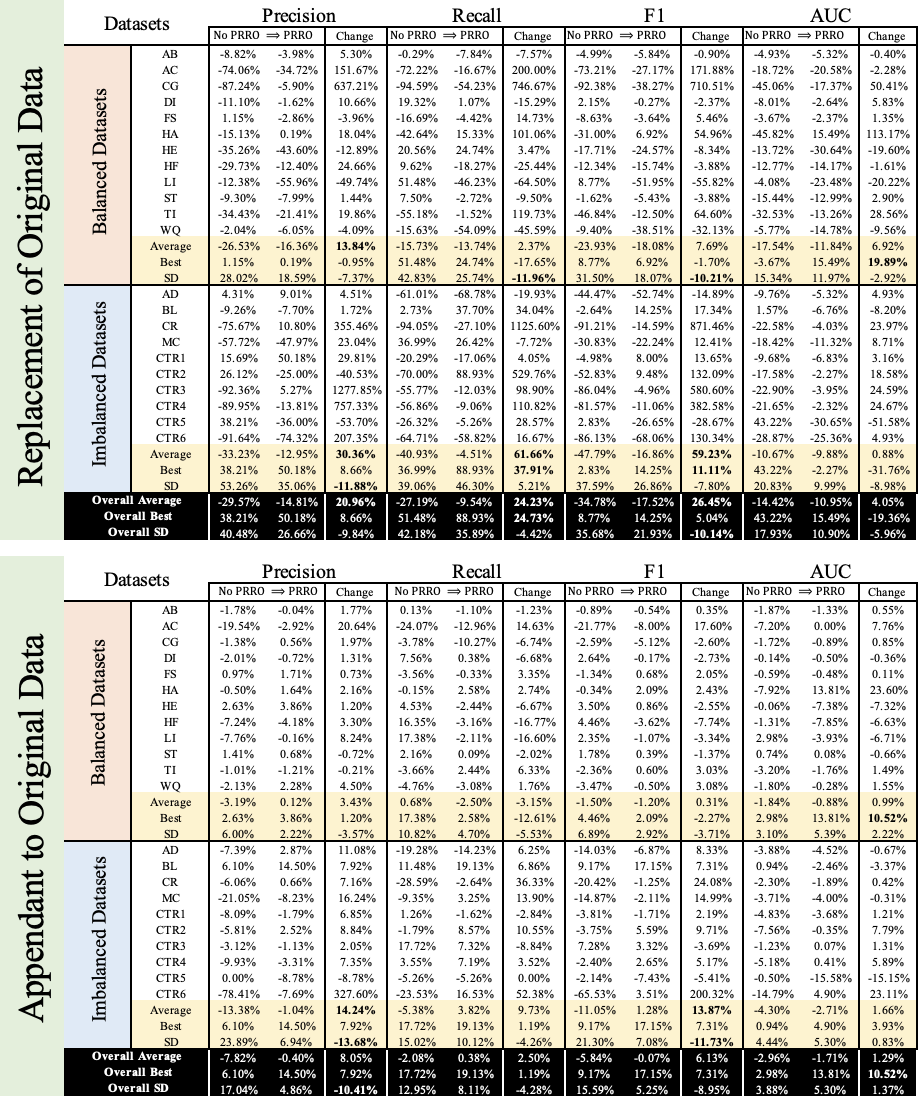} 
\caption{The graph shows SL utility comparison between original and synthetic data. Bold numbers indicate that the change after implementing PRRO pipeline that results in a more than 10\% improvement for illustrative purpose. Synthetic data utility is evaluated in two aspects: (1) replacement: whether the synthetic data alone can perform as well as the original data, and (2) appendant: whether the addition of synthetic data can further enhance prediction performance of the original data. It is notable that PRRO implementation leads to a significant increase in both the average and best cases of classification metrics but a decrease in the standard deviation, summarizing a consistent and significant improvement in synthetic data utility after implementing PRRO.}  
\label{fig: result-summary}
\end{figure*}

Three experimental findings, illustrated by Table \ref{tab: pruning-results} and Fig. \ref{fig: result-summary}, highlight a consistent improvement of synthetic data utility after implementing the PRRO pipeline. In Fig. \ref{fig: result-summary}, the two values before and after using PRRO are absolute utility change compared to the baseline as defined in Def. \ref{def: synthetic-data-utility}, where the change shows the relative change of synthetic data utility after implementing the PRRO pipeline. 
\vspace{-2mm}

\subsubsection{\textbf{Over 20\% improvement in all classification accuracy metrics on synthetic replacement using PRRO}} \label{subsub: improvement1}
The improvement in synthetic data utility in the replacement case is exceptionally notable after implementing the PRRO pipeline. The classification accuracy using original data serves as the upper bound of synthetic model utility in replacement scenario, given the fact that synthetic data is synthesized from the original data and hence is a derivative of the original data. Based on this consideration, PRRO pipeline allows a closer resemblance to the upper bound. \textbf{On average, it improves all accuracy-related metrics for more than 20\%, including a 20.96\%, 24.23\% and 26.45\% improvement in the precision, recall and F1 respectively.} With a 6.92\% increase in the AUC score, it confirms that the conditional relationship between the predictor and features becomes more visible after implementing PRRO, as the SL model is able to make predictions with increased confidence. The consistency of outperformance is further supported by a drop of standard deviation for all metrics quantified by a 9.84\%, 4.42\% and 10.14\% drop in the precision, recall and F1 metrics respectively, implying a closer distance to the upper bound. These increases in all accuracy metrics and drops of standard deviation summarize a consistent outperformance of synthetic data utility after PRRO implementation. This uniform improvement points to the value of the Column Conditional ReOrdering module, as the significance lies between setup before and after the pipeline implementation primarily differentiated by the use of the reordering module. The values of the Signal-based Data Pruning module is illustrated in the more material improvement over the imbalanced datasets, with increase in accuracy being over 30\%, as the module is specifically designed to process imbalanced datasets. Further benefits of the Signal-based Data Pruning module are discussed in Sec. \ref{subsub: improvement3}, where the detailed label distribution of datasets is also explored. 

\subsubsection{\textbf{A 6.13\% improvement in F1 score on synthetic appendant using PRRO}} \label{subsub: improvement2}
\begin{figure}[H]
    \centering
    \includegraphics[width=0.95\linewidth]{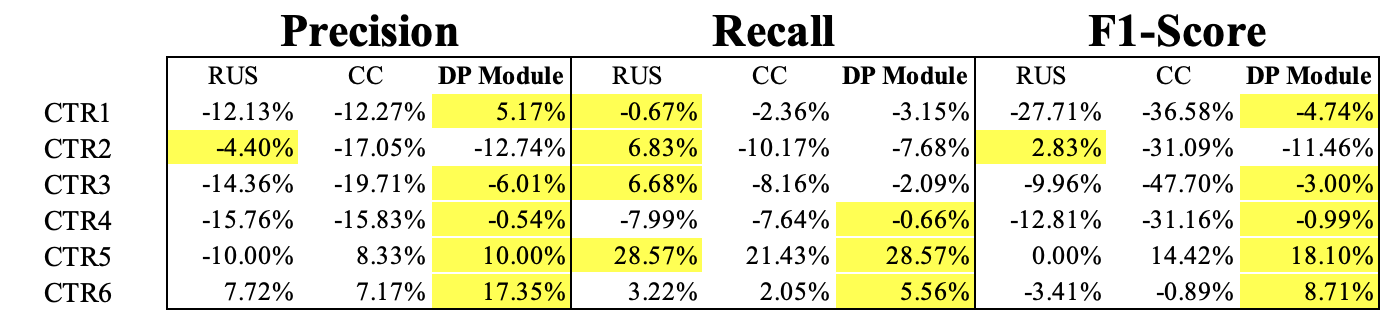}
    \caption{Conducting the same experiments with two undersampling techniques, i.e. Random Undersampling (`RUS') and Cluster Centrois (`CC') shows that the Data Pruning Module (Sec. \ref{sub: signal-subgrouping}) can effectively outperform existing undersampling techniques in terms of synthesizing data with high supervised learning utility. }
    \label{fig:data-pruning-module-comparison}
\end{figure}

In terms of the synthetic data utility under synthetic appendant case, an improvement of synthetic data quality is still observed over all evaluation metrics, despite being not as significant as those in the synthetic replacement case. It is observed that the implementation of PRRO still leads to 6.13\% improvement, hence showing the value of the Colunn Conditional ReOrdering module. However, unlike the replacement case, the baseline with original data only should serve as a soft lower bound, as synthetic data is added on top of the original data. With a slight 0.07\% drop on the F1 score, the addition of synthetic data does not improve overall data quality in fact. Contradictorily, the degradation of classification performance indicates SL model confusion by the addition of synthetic data. Nevertheless, by analyzing further on the imbalanced datasets, we can observe a more concrete synthetic data utility. To begin with, there is a 13.87\% improvement of F1 score after PRRO implementation, and more importantly, 1.28\% increase compared to the baseline, implying that PRRO can assist synthetic data in improving supervised learning utility in imbalanced scenario. This can be reasoned with the usage of Signal-based Data Pruning module. By adding synthetic minority data (Sec. \ref{sub: signal-subgrouping}) to the original dataset to distort the original class distribution (Table \ref{tab: pruning-results}), the SL model can weight more heavily on minority classes with an increasing exposure of the minorities. The significant drop in standard deviation of 11.73\% further supports the module consistency under different datasets. A notable observation is that the 1.28\% improvement over the baseline is contributed by a 3.82\% increase in recall and a 1.04\% drop in precision, which is precisely the impact of the Signal-based Data Pruning module. With a distorted data distribution weighting the minority classes more heavily, this leads to a higher probability of false positive over false negative, i.e. the precision-recall trade-off. However, it is generally accepted that, under an imbalanced dataset scenario, the benefit of improving recall (identifying more instances of the minority class) typically outweighs the drawbacks of a potential increase in false positives (precision). The benefit is reflected by the net improvement of F1 score. Nevertheless, there are still cases where precision is more important than recall even at highly imbalanced datasets \cite{precision-recall}, such as amputation analysis, where the cost of false positive immensely outweighs that of false positive. All in all, the exact balance between precision and recall varies across different contexts and datasets. Moreover, we compare the efficacy of the Data Pruning module against various undersampling methods including Random Undersampling and Cluster Centroids \cite{rus, cc}, where Fig. \ref{fig:data-pruning-module-comparison} shows that synthetic data generated under our Data Pruning module has a consistent outperformance on classification F1-Score, as well as precision. 

\subsubsection{\textbf{43\% improvement in supervised learning utility after data pruning}} \label{subsub: improvement3} \vspace{-2mm}
\begin{table}[H]
    \centering
    \begin{tabular}{|l|l|l|l|l|}
    \hline
        Datasets & \multicolumn{2}{l|}{Without Pruning} & \multicolumn{2}{l|}{With Pruning} \\ \hline
        ~ & Original & Synthetic & Original & Synthetic  \\ \hline
        CDR1 & 0.9777 \% & 0.0889\% & 4.1448\% & 0.4306\%  \\ \hline
        CDR2 & 1.6568\% & \textcolor{red}{0.0200\%} & 9.9960\% & 14.9338\%   \\ \hline
        CDR3 & 1.2662\% & \textcolor{red}{0.0015\%} & 5.0711\% & 0.7652\%   \\ \hline
        CDR4 & 2.1325\% & \textcolor{red}{0.0018\%} & 9.1426\% & 6.0823\%   \\ \hline
        CDR5 & 4.0541\% & 5.4054\% & 26.3158\% & 31.5789\%   \\ \hline
        CDR6 & 1.6252\% & 0.1912\% & 13.0769\% & 8.4615\%  \\ \hline
        Average & 1.9521\% & 0.9515\% & 11.2912\% & 10.3754\% \\ \hline
    \end{tabular}
    \caption{Pruning does not only improve the positive rate for subsequent data analytics ease, but also allows the generator to generate minority class at a rate more similar to the original data. Colored synthetic cases had a 0\% positive rate in fact, which made subsequent supervised learning utility evaluation impossible. Therefore, one random point was arbitrarily chosen to be positive, contributing to the non-zero positive rate.} \vspace{-5mm}
    \label{tab: pruning-results}
\end{table}
The Signal-based Data Pruning module specifically improves synthetic data utility of imbalanced datasets. When studying the positive rate of the extremely imbalanced subdatasets from \cite{ctr_dataset}, Table \ref{tab: pruning-results} shows discount in minority distribution in synthetic data compared to original data. The module significantly reduces the rate of discount after pruning. This does not only quantify the improvement of class distribution similarity, but also contributes to synthetic data utility improvement in Fig. \ref{fig: result-summary}. Precisely, the positive rate is observed to be extremely imbalanced where none of the positive rate is above 5\% and the average is as low as 1.95\%. \textbf{Before data pruning}, the synthetic positive rate is \textbf{discounted by $51\%$} from 1.95\% to 0.95\%, exaggerated by the generator. \textbf{After pruning}, with the positive rate improving to 11.29\%, \textbf{a significantly smaller 8\% discount rate} from 11.29\% to 10.38\% can be observed. This accounts for \textbf{a 43\% improvement} in similarity of class distribution between original and synthetic data. This drop in discount rate highlights more similar class distribution between original and synthetic data, hence underscoring the value of signal-based data pruning method in handling highly imbalanced datasets.\\

\section{Summary and Future Work} \label{sec: future work}
In this work, we introduce the Pruning and Reordering PRRO pipeline, a novel data-centric framework designed to enhance the utility of synthetic data generated by tabular data generators. Our approach systematically addresses key challenges associated with generators that exaggerate imbalance class distribution and overlook necessary conditional data relationships of supervised learning models, tackling the challenges with the Signal-based Data Pruning and Column Conditional ReOrdering modules respectively. Our experimental results across 22 diverse datasets show that synthetic data utility improvement after implementing the PRRO pipeline. Precisely, the Column Conditional ReOrdering module allows a generalized improvement of synthetic data utility, after addressing the generator's hurdle in overlooking predictor-features conditional relationship. Furthermore, the synthetic data utility is further enhanced in the case of imbalanced datasets thanks to the Signal-based Data Pruning module, which addresses the imbalance dataset exaggeration within generators. The PRRO pipeline allows generator to synthesize more original-data alike synthetic data to accomplish an improved efficacy in the synthetic replacement task. While synthetic data utility improvement is more material in synthetic replacement scenario, the PRRO also contributes to a betterment of synthetic appendant to minority class, which allows SL model trained under the newly combined dataset to capture patterns of minority class more easily. All in all, the PRRO pipeline bridges the critical gaps between state-of-the-art generators and SL models, streamlining the 'generate and predict' process. \\
\\
Looking ahead, we see possible utility improvement by further aligning the data structure between tabular data and LLMs. Other than the textual encoding technique that converts each row observation into a sentence \cite{borisov2023language, solatorio2023realtabformergeneratingrealisticrelational}, there are also other works seeking to effectively encode tables for LLM fine-tuning and training \cite{li2024graphneuralnetworkstabular, su2024tablegpt2largemultimodalmodel, fang2024largelanguagemodelsllmstabular}. By developing a model structure that can effectively study both column-wise and row-wise relationships across a table, the application of LLM can be immensely extended in the area of tabular data. Aside from developing a tabular encoder that fully complies to tabular data, we can also explore the betterment of tabular modeling with existing LLMs, by leveraging the corresponding position bias to allocate corresponding weights to different column relationships. There have been works exploring the mitigation and alignment of position bias in LLM \cite{yu2024mitigatepositionbiaslarge, li2024splitmergealigningposition}, motivating an further step in handling LLM position bias - utilizing the bias to improve model performance. \\
\\
Specifically, we propose an extension of the Column Conditional ReOrdering algorithm to adapt to potential multicollinearity problem in real-life datasets. Given the column-by-column conditional relationship modeling of the tabular generators, we hypothesize leveraging this modeling structure by arranging the features ordered by importance to the predictor, so that columns of higher importance can be further emphasized by the column. Algo. \ref{algo: optimal-feature} proposes reordering the features in terms of the importance, e.g. quantified by permutation importance score from \cite{featureimportance}. Next, given the column-by-column conditional relationship adopted by generators, we hypothesize that columns that are modeled in the beginning have less probability constraints, compared to the generation of later columns which have to follow the conditional density of previous features, thereby leading to a higher degree of randomness. Therefore, by placing the more important features at the end, this guides the generator in modeling and synthesizing important features more carefully (with more constraints), preserving data relationship of higher importance and hence enhancing synthetic data utility. With future works hypothesizing improvement in generator models and training data quality, we are devoted to continue advancement in the area of trustworthy data synthesis to promote safer and more accessible data analytics.

\bibliographystyle{ACM-Reference-Format}
\bibliography{sample-base}

\appendix
\section{Appendix}
\subsection{Textual Encoding Transformation for LLM-based Tabular Synthesis} \label{sub: textual-encoding}
Textual encoding transformation $T(\cdot)$ is a method proposed in \cite{borisov2023language} that transform tabular observations into natural language for LLM processing. Given a dataset with $n$ observations and $k$ features ($\mathcal{D}=\{x_{i1},x_{i2},...,x_{ik}\}_{i=1}^{n}$, each observation is converted into a sentence as follows: 'Column 1: $x_{i1}$, Column 2: $x_{i2}$, $...$, Column $k$: $x_{ik}$', using comma (,) and colon (:) for separation between columns and within column name and column value. The sentences are then used to fine-tune the LLM to generate texts that mimic the behavior of original texts. These semantic-preserving synthetic sentences are inversely transformed back to the tabular form \cite{borisov2023language}.

\subsection{Detailed summary for the datasets used in Sec. \ref{sec: experiment} Experiment}
\begin{table}[H]
    \centering
    \resizebox{\linewidth}{!}{
    \begin{tabular}{ccccccc}
        \hline
       Abbr.  & Name & \# Train (Holdout) & \# Test & \# Num & \# Cat & Positive Rate \\
       \hline 
        AB & Abalone & 1,670 & 836 & 7 & 1 & 49.82\% \\
        AC & Acute Inflammation & 48 & 24 & 1 & 7 & 39.58\% \\
        AD & Adult Income & 14,652 & 7,327 & 5 & 8 & 23.93\%\\
        BL & Blood Donation & 229 & 150 & 3 & 0 & 23.75\%\\
        CG & Congress & 174 & 87 & 0 & 16 & 38.51\%\\
        CR & Credit & 9,600 & 4,800 & 14 & 9 & 22.18\%\\
        DI & Diabetes & 28,276 & 14,139 & 3 & 18 & 50.00\%\\
        FS & Flight Satisfaction & 41,561 & 20,781 & 4 & 18 & 43.33\%\\
        HA & Haberman & 122 & 62 & 3 & 0 & 73.77\%\\
        HE & Heart Disease & 104 & 53 & 5 & 5 & 37.5\%\\
        HF & Heart Failure & 119 & 60 & 7 & 5 & 31.93\%\\
        LI & Liver Disease & 233 & 177 & 9 & 1 & 28.76\%\\
        MC & Marketing Campaign & 896 & 448 & 17 & 10 & 14.96\%\\ 
        ST & Student Performance & 400 & 200 & 2 & 5 & 51.5\%\\
        TI & Titanic & 356 & 179 & 2 & 7 & 38.48\%\\
        WQ & Wine Quality & 457 & 229 & 11 & 0 & 54.27\%\\
        CTR1 & CTR Group 1 & 31,503 & 15,752 & 4 & 19 & 0.98\%\\
        CTR2 & CTR Group 2 & 15,029 & 7,515 & 4 & 19 & 1.66\%\\
        CTR3 & CTR Group 3 & 65,946 & 32,974 & 4 & 19 & 1.27\%\\
        CTR4 & CTR Group 4 & 55,756 & 27,878 & 4 & 19 & 2.13\%\\
        CTR5 & CTR Group 5 & 148 & 75 & 4 & 19 & 4\% \\
        CTR6 & CTR Group 6 & 1,041 & 524 & 4 & 19 & 1.63\%\\
        \hline
    \end{tabular}}
    \caption{Summary of the datasets used for the experiment to test the PRRO pipeline}
    \label{tab: data-summary}
\end{table}

\subsection{Extension of Column Conditional ReOrdering algorithm for multicollinearity adaptation}
\begin{algorithm}[H]
\caption{Finding optimal order for generator}\label{algo: optimal-feature}
\begin{algorithmic}[1]
\STATE \textbf{Input:} A training and a validation dataset with $p$ features: $D_{T/V}=\{(\mathbf{x}_{T/V,i},y_{T/V,i})\}_{i=1}^{k_{T/V}}$, where $\mathbf{x}_{T/V,i}=(x^{(1)}_{T/V,i}, \dots, x^{(p)}_{T/V,i})$.
\STATE \textbf{Output:} Optimal feature order that maximizes utility of generator $\mathcal{G}_{A}$.

\STATE Initialize generator $\mathcal{G}_{A}$ predictive model $h$, $err \gets \infty$;
\STATE $X \gets \mathbf{x}_{T}$;
\STATE $y \gets y_{T}$;
\STATE Train $h$ with $X$ and $y$;
\STATE Rank $X$ according to permutation importance score to get the ranking $r$;
\STATE \textbf{Return} $r$;
\STATE $X \gets X[r]$ \COMMENT{Reorder $X$ according to the ranking};
\STATE Train $\mathcal{G}_{A}$ by concatenating $X$ and $y$;
\STATE Synthesize $\tilde{X}$, $\tilde{y}$;
\STATE Measure utility between $\{\tilde{X},\tilde{y}\}$ and $\{X, y\}$ using validation dataset $D_{V}$;
\end{algorithmic}
\end{algorithm}

\end{document}